%% file: naacl_2025.tex
\pdfoutput=1

\documentclass[11pt]{article}

\usepackage[final]{acl}

\usepackage{times}
\usepackage{latexsym}

\usepackage[T1]{fontenc}

\usepackage[utf8]{inputenc}

\usepackage{microtype}

\usepackage{inconsolata}

\usepackage{graphicx}
\usepackage[normalem]{ulem}
\usepackage{multirow}

\usepackage{booktabs}

\title{
Richer Output for Richer Countries: 
Uncovering Geographical 
\\ Disparities in Generated  Stories and Travel Recommendations
}

\author{Kirti Bhagat \\
    Indian Institute of Science \\
    Bengaluru, KA, India \\
    \texttt{kirtibhagat@iisc.ac.in} \\\And
    Kinshuk Vasisht \\
    Indian Institute of Science \\
    Bengaluru, KA, India \\
    \texttt{kinshukv@iisc.ac.in} \\ \And
    Danish Pruthi \\
    Indian Institute of Science \\
    Bengaluru, KA, India \\
    \texttt{danishp@iisc.ac.in} \\}

\begin{document}
\maketitle
\begin{abstract}
\input{sections/abstract}

\end{abstract}

\input{sections/introduction}

\input{sections/relatedWork}

\input{sections/approach}

\input{sections/results}

\input{sections/limitations}

\input{sections/conclusion}

\section*{Acknowledgements}
\input{sections/acknowledgements}

\bibliography{naacl_2025}

\input{sections/appendix}

\end{document}

%% file: sections/abstract.tex
While 
a large body of work inspects language models for biases concerning gender, race, occupation and religion, 
biases of geographical nature are relatively less explored. 
Some recent studies
benchmark the degree to which large language models 
encode geospatial knowledge. 
However, the impact of  
the encoded geographical knowledge (or lack thereof) 
on real-world applications 
has not been documented.
In this work, 
we examine 
large language models 
for two common scenarios that require 
geographical knowledge: (a) travel recommendations and (b) geo-anchored story generation. 
Specifically, we study 
five popular language models, 
and across about $100$K travel requests, and $200$K story generations, 
we observe that
travel recommendations corresponding to poorer countries 
are \emph{less unique} with fewer location references, 
and stories from these regions more often convey emotions of hardship and sadness compared to those from wealthier nations.
\footnote{The data and code is available at \url{https://github.com/FLAIR-IISc/richer-countries-have-richer-output}.}

%% file: sections/introduction.tex
\section{Introduction}

\begin{figure*}[t]
\centering
\includegraphics[width=\textwidth]{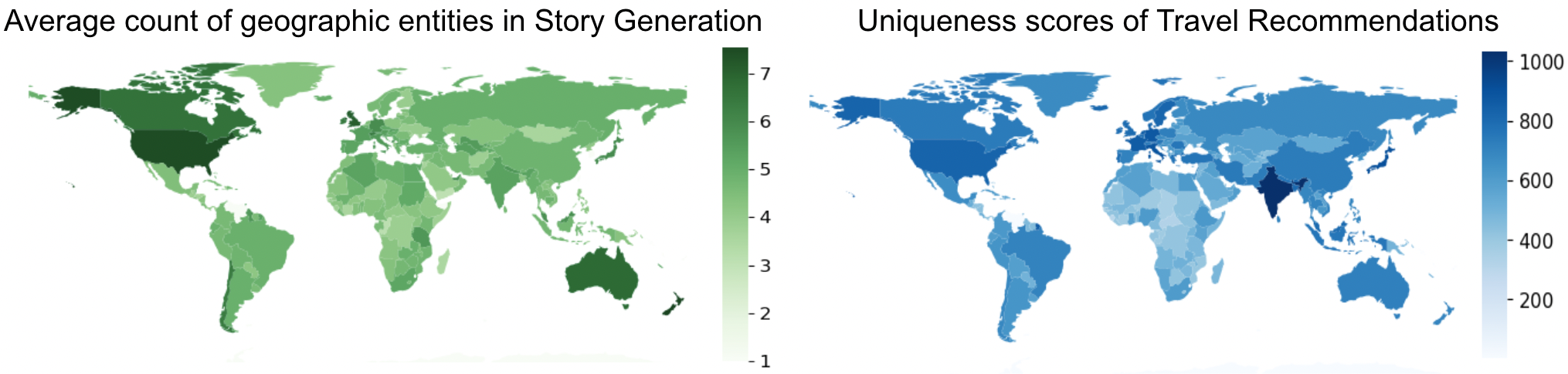}
\caption {World map with country-wise analysis of responses generated by GPT-4. Left: Average count of geographical entities mentioned in generated stories (correlated with the GDP per capita with Pearson $r$ = $0.5$). Right: Uniqueness scores for travel recommendations (Pearson $r$ = $0.4$ with GDP  per capita). 
}
\label{fig:world_maps}
\end{figure*}

Given the excitement around large language models, users 
resort to these models for a 
diverse range of applications~\citep{brown2020language, touvron2023llama}. 
Based on our analysis 
of ShareGPT,\footnote{\url{https://sharegpt.com/}} a  
collection of user interactions with ChatGPT, 
$1.7$\% of queries 
are about travel recommendations,  
whereas $1.5$\% concern story generation. 
Such use cases make one wonder whether
the generated travel itinerary for Mumbai is just as informative 
compared to New York City?
Or is a generated story of 
a girl growing up 
in  Nairobi
just as relatable 
compared to another story based in Seattle?
For these applications 
to be broadly useful, it is important that there are no (or few) geographical disparities.

Some recent works aim  
to benchmark the extent of 
geographical knowledge encoded 
in large language models \citep{bhandari2023large, manvi2023geollm, moayeri2024worldbench}.
Interestingly, a recent study finds that 
language models accurately predict 
global facts such as population and rainfall 
for different geo-locations, 
however, their predictions 
on
sensitive topics 
such as attractiveness or morality
are, problematically, biased against 
areas with poorer socioeconomic conditions \citep{manvi2024large}.
Similar in spirit, 
our work aims 
to quantitatively 
study model responses 
for two real-world scenarios that require 
geographical knowledge.

In this work,
we analyze
over $300$K responses from $5$ language models,
corresponding to requests 
for travel recommendations 
and geo-anchored story generations. 
These requests span over $150$K
locations across the globe.
We quantify the 
informativeness and uniqueness of 
model responses, 
in addition to the emotions they express. 
We then compare these quantities  
with the socioeconomic indicators 
of different locations.

Through our analysis, we 
uncover several geographical disparities, 
finding that 
outputs for wealthier countries 
are more unique and include 
more geographical entities (Figure~\ref{fig:world_maps}). For Sub-Saharan African countries, we notice considerably less unique outputs compared to the North American region, with the average difference being about $40$\% across all models.
Further, 
stories about poorer countries 
express considerably more hardship 
and sadness, with $60$\% fewer narratives depicting hardship for high-income countries compared to low-income countries.

Despite many large corporations 
claiming to be egalitarian, e.g., OpenAI aims to develop intelligence that \emph{benefits all of humanity} \citep{OpenAIAbout}, 
many findings, including ones from this study, demonstrate how  current models 
perpetuate western hegemony in the generated content \cite{schwobel2023geographical, bender2021dangers}, and 
more serious effort is needed to 
ensure that models serve 
the diverse population across the globe.

%% file: sections/relatedWork.tex
\section{Related Work}
\label{sec:related_work}

\textbf{Geographical Bias in Language Technologies.}
Language models can generate disproportionate and prejudiced representations of marginalized groups \citep{navigli2023biases}. 
Significant efforts quantify these biases across various dimensions, including gender \citep{sheng2019woman}, race \citep{omiye2023large}, culture \citep{wang2023not}, religion \citep{abid2021persistent}, and more.
Recent studies also highlight `geographical biases' in LMs:
one such study finds that when models are instructed to rank countries
in terms of topics such as work ethic,
intelligence and attractiveness, they
undervalue areas with lower socio-economic status \citep{manvi2024large}.
An analysis of models over WorldBench, a benchmark to assess factual recall in LLMs over country-wide data from the World Bank,
reveals higher error rates for countries with lower income levels \citep{moayeri2024worldbench}.
Global-Liar, a dataset by \citet{mirza2024global}, benchmarks the accuracy of models in fact-checking claims from six global regions and underscores the disadvantages faced by regions in the Global South.
Despite these efforts to highlight bias in geographical factual predictions, there is a noticeable lack of research addressing biases in real-world applications of geographical knowledge. Our work addresses this gap by examining biases across practical applications of story generation and travel recommendation.
There exists prior work 
on travel planning \cite{xie2024travelplannerbenchmarkrealworldplanning}, and a large literature on story generation \cite{zhao2023story}, but to the best of our knowledge, these works do not examine 
geographical disparities.

%% file: sections/approach.tex
\section{Approach}

\input{tables/metrics.corr.tex}

Below, we briefly 
describe 
our approach to quantitatively 
examine 
the outputs from different language models 
for two tasks: generating travel recommendations and geo-anchored stories.

\subsection{Experimental Setup}

\paragraph{Examined Locations.} 
To capture the global variation in LLM performance, it is crucial to incorporate a diverse array of geographical regions, covering cities, towns, and villages, in our analysis.
For this purpose, we use Geonames\footnote{\url{https://www.geonames.org/}}, a community-driven geographical database, comprising an extensive list of location names. 
We capture all global locations with inhabitants exceeding $1000$, to achieve a finer granularity and a more nuanced understanding of global disparities. This surpasses 
most prior 
work 
that largely study geographical biases with only country-level information (\S\ref{sec:related_work}).
At the time of our study, GeoNames contained about $150$K locations.
To make the analysis computationally feasible, we randomly sample, with replacement, up to $25$ locations per country from this larger population totalling about $4,000$ locations for $190$ countries. For every attribute we quantify, we report averages across $3$ such random samples (with different seeds) to infer patterns across all locations.

\paragraph{Input Prompts.}  
We manually curate a set of prompt templates with location placeholders, which are later populated with the sampled locations. For example, "Write a story of a family from [Location]" is a prompt for story generation. We intentionally keep the prompts simple, avoiding additional variables to ensure a fair comparison and isolate the influence of location on the model's response.
For each location, we randomly choose 6 templates (4 for story generation and 2 for travel recommendation) and fill the location slot. Such slot-filling approaches to generate inputs are commonly used in the literature \citep{chang2023prompt}.
While queries related to travel recommendations naturally incorporate a geographical aspect, for story generation, we specifically design \emph{system prompts} to elicit accurate and descriptive geographical details. We share the list of prompt templates in \hyperref[sec:Appendix Prompt Templates]{Appendix \ref{sec:Appendix Prompt Templates}}.

\paragraph{Models.} We perform our evaluation on five widely used and 
capable models: GPT-$4$ (\citealp{achiam2023gpt}), Mistral 7B (\citealp{jiang2023mistral}), Mixtral $8\times7$B (\citealp{jiang2024mixtral}), LLaMa $3$ $8$B, and LLaMa $3$ $70$B (\citealp{dubey2024llama}), to help us cover a broad range of characteristics, including size, training methods, and availability. We set the temperature to $0.7$ to achieve a balance between creativity and coherence.

\subsection{Evaluation}

\begin{figure*}[t]
  \includegraphics[width=0.48\linewidth]{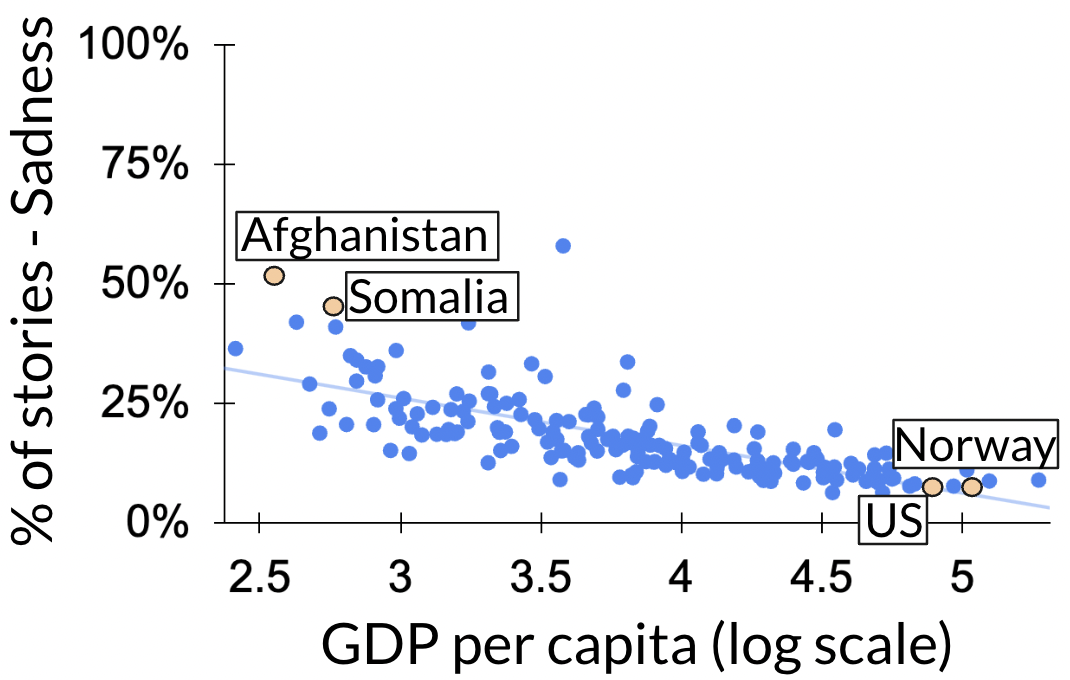} \hfill
  \includegraphics[width=0.48\linewidth]{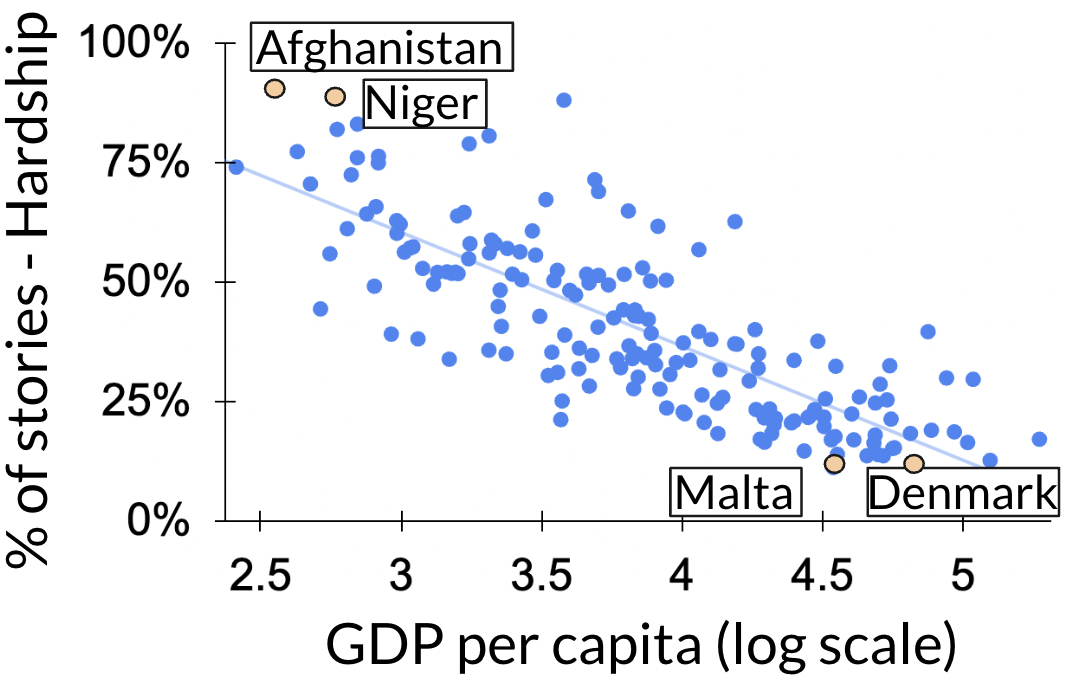} \hfill
  \caption {Percentage of stories generated by GPT-4 depicting the emotions of sadness (left, Pearson $r$ = $-0.45$) and hardship (right, Pearson $r$ = $-0.54$) for each country vs. GDP per capita.
}
\label{fig:emotion_trend}
\end{figure*}

For geo-anchored applications, such as travel recommendations and story generation,  
there is often
no definitive right or wrong response. 
One possible approach is to find local participants 
from every geographical region, 
and request them to qualitatively 
evaluate each response. Past work has noted that it is 
challenging to find such participants \citep{basu2023inspecting}. 
Instead, we 
quantify few attributes that we believe 
contribute to the quality of the response, such as uniqueness and informativeness. %
We briefly describe these attributes:

\paragraph{Uniqueness.} 
Every location presents a blend of historical, cultural, geographical, and environmental factors, contributing to its unique identity. To reflect this, models need to be aware of these distinctive aspects for different regions. We use the uniqueness measure to capture how distinct the subjective responses for a given location are compared to others. We make a slight modification to the TF-IDF metric to calculate the average rarity of words generated in the response of a location compared to other locations. Terms which score high as per our metric, anecdotally, reflect regional artifacts (e.g., ``dosa'' for India). 
Additionally, we 
also compute 
the Type-Token Ratio (TTR),
the ratio of unique words to total words,
which captures the lexical 
diversity of these responses. 
Past efforts have effectively used the TTR metric to measure richness in vocabulary in various contexts \citep{balestrucci2024m, morris2023using}.
We provide further details of the uniqueness metric in \hyperref[sec:Appendix Uniqueness Calculation]{Appendix \ref{sec:Appendix Uniqueness Calculation}}.

\paragraph{Informativeness.}
 It is challenging to 
 quantify the overall informativeness
 of a response, as a proxy, we 
 compute the number of 
 geographical entities 
 present in model responses.  
 Both the examined applications 
 solicit geographical details, 
 and therefore we tag
the responses using Spacy NER tagger to extract GPE (Geopolitical entities), LOC (Non-GPE locations) and FAC (buildings, airports, etc.).
 We also attempt to quantify 
 the factual correctness, or utility, of 
 recommendations
 by estimating the proximity 
 of the extracted locations to the city in question. 
 However, quantifying this turns out to be quite challenging (see \S\ref{sec:limitations} for details).

\paragraph{Emotions.} 
Stories are rife with emotions. We study the variation in emotions expressed in model generated stories.
Recent research has demonstrated that GPT-$4$ can serve as an effective emotion annotator \citep{niu2024text}, noting that human evaluators often favour emotion annotations by GPT-$4$ over those provided by human annotators. We apply a similar method for emotion recognition in generated stories by using GPT-4 to identify the emotions of joy, hardship and sadness expressed.

%% file: tables/metrics.corr.tex
\begin{table*}
\small
\centering
\addtolength{\tabcolsep}{-0.3em}
\begin{tabular}{@{}llcc|cc|cc|cc|cc|@{}}
\toprule
\multicolumn{1}{l}{\textbf{Scenario}} & \multicolumn{1}{l}{\textbf{Attribute}} & \multicolumn{2}{c}{\textbf{GPT4}} & \multicolumn{2}{c}{{\textbf{Mistral 7B}}} & \multicolumn{2}{c}{\textbf{Mixtral 8x7B}} & \multicolumn{2}{c}{\textbf{LLaMa3 8B}} & \multicolumn{2}{c}{\textbf{LLaMa3 70B}} \\ \midrule

\multicolumn{1}{l}{\multirow{6}{*}{Travel}}&  & \multicolumn{1}{|c}{GDP} & Freq & \multicolumn{1}{|c}{GDP} & Freq & \multicolumn{1}{|c}{GDP} & Freq & \multicolumn{1}{|c}{GDP}  & Freq & \multicolumn{1}{|c}{GDP} & Freq \\ \midrule

\multicolumn{1}{l}{\multirow{5}{*}{Rec.}}  & Uniqueness & \multicolumn{1}{|c}{$~~0.39^*$} & $~~0.47^*$ &  \multicolumn{1}{|c}{$~~0.32^*$} & $~~0.48^*$ & \multicolumn{1}{|c}{$~~0.35^*$} & $~~0.48^*$ & \multicolumn{1}{|c}{$~~0.27^*$} & $~~0.53^*$ & \multicolumn{1}{|c}{$~~0.31^*$} & $~~0.45^*$ \\
 
 & \# Geo-entities & \multicolumn{1}{|c}{$~~0.20^*$} & $~~0.14~~$ & \multicolumn{1}{|c}{$~~0.30^*$} & $~~0.16^*$ & \multicolumn{1}{|c}{$~~0.16^*$} & $~~0.19^*$ & \multicolumn{1}{|c}{$~~0.12~~$} & $~~0.22^*$ & \multicolumn{1}{|c}{$~~0.42^*$} & $~~0.30^*$ \\

 & {TTR} & \multicolumn{1}{|c}{$-0.26^*$} & $~~0.09~~$ & \multicolumn{1}{|c}{$-0.08~~$} & $~~0.03~~$ & \multicolumn{1}{|c}{$-0.20^*$} & $~~0.05~~$ & \multicolumn{1}{|c}{$-0.19^*$} & $-0.05~~$ & \multicolumn{1}{|c}{$-0.23^*$} & $~~0.00$ \\
 
 & Absence of Info & \multicolumn{1}{|c}{$-0.37^*$} & $-0.04~~$ & \multicolumn{1}{|c}{$-0.09~~$} & $~~0.09~~$ & \multicolumn{1}{|c}{$-0.25^*$} & $-0.02~~$ & \multicolumn{1}{|c}{---} & --- & \multicolumn{1}{|c}{---} & --- \\ \midrule

\multicolumn{1}{l}{\multirow{4}{*}{Story}}  & Uniqueness & \multicolumn{1}{|c}{$~~0.25^*$} & $~~0.41^*$ & \multicolumn{1}{|c}{$~~0.08~~$} & $~~0.27^*$ & \multicolumn{1}{|c}{$~~0.15^*$} & $~~0.32^*$ & \multicolumn{1}{|c}{$~~0.28^*$} & $~~0.39^*$ & \multicolumn{1}{|c}{$~~0.26^*$} & $~~0.38^*$ \\

\multicolumn{1}{l}{\multirow{4}{*}{Gen.}} & \# Geo-entities & \multicolumn{1}{|c}{$~~0.49^*$} & $~~0.37^*$ & \multicolumn{1}{|c}{$~~0.35^*$} & $~~0.36^*$ & \multicolumn{1}{|c}{$~~0.48^*$} & $~~0.41^*$ & \multicolumn{1}{|c}{$~~0.46^*$} & $~~0.44^*$ & \multicolumn{1}{c}{$~~0.49^*$} & $~~0.43^*$ \\

  & TTR & \multicolumn{1}{|c}{$~~0.14~~$} & $~~0.08~~$ & \multicolumn{1}{|c}{$~~0.39^*$} & $~~0.12~~$ & \multicolumn{1}{c}{$~~0.28^*$} & $~~0.08~~$ & \multicolumn{1}{c}{$~~0.26^*$} & $~~0.14~~$ & \multicolumn{1}{c}{$~~0.52^*$} & $~~0.22^*$ \\
  
 & Hardship & \multicolumn{1}{|c}{$-0.54^*$} & $-0.20^*$ & \multicolumn{1}{|c}{$-0.38^*$} & $-0.13~~$ & \multicolumn{1}{c}{$-0.50^*$} & $-0.20^*$ & \multicolumn{1}{c}{$-0.42^*$} & $-0.17^*$ & \multicolumn{1}{c}{$-0.46^*$} & $-0.16^*$ \\
 
 & Sadness & \multicolumn{1}{|c}{$-0.45^*$} & $-0.17^*$ & \multicolumn{1}{|c}{$-0.28^*$} & $-0.12~~$ & \multicolumn{1}{c}{$-0.34^*$} & $-0.13~~$ & \multicolumn{1}{c}{$-0.32^*$} & $-0.12~~$ & \multicolumn{1}{c}{$-0.30^*$} & $-0.10~~$ \\ \bottomrule
 
\end{tabular}
\caption{Pearson's correlation coefficients depicting the relationship between various attributes (averaged for each country) with the GDP per capita
(denoted as \textbf{GDP}) and the frequency of country mentions in the PILE dataset (\textbf{Freq}) for generating travel recommendations and stories. $*$ denotes values with p-value $<0.05$. 
}
\label{tbl:results_GDP_coorelations}
\end{table*}

%% file: sections/results.tex
\section{Results}

We aggregate the (estimated) city-level attributes for every country to observe trends across different countries. Interestingly, the uniqueness scores for India, Italy, Japan, and the United States consistently rank among the highest across all models in both applications. Comparing regions, we observe that travel recommendations for the Sub-Saharan African region are considerably less unique than those for the North American region, with alarming differences of $43$\% for GPT-4, $42$\% for LLaMa $3$ $8$B, $39$\% for LLaMa $3$ $70$B, $37$\% for Mistral $7$B,
and $38$\% for Mixtral $8\times7$B .

A similar trend is observed in the informativeness of generated stories, wherein 
most models 
generate a large number of geographical entities for the United States and the United Kingdom.
Problematically, the responses for the North American region include around double the number of location mentions in stories compared to those from the Sub-Saharan African region across all models.

Studying the emotions presented in the generated stories, we notice that  
while only $20$\% of stories generated for the North American region depict hardship, this figure rises to $47$\% for the Sub-Saharan African region.
Interestingly, we find that an overwhelming fraction of stories, about $99\%$, express some form of joy, suggesting an overall \textbf{positivity bias} in model-generated stories.

\paragraph{Relationship with GDP per capita.}
We present the Pearson correlation coefficients for all attributes aggregated by country with the country's GDP per capita for different models in Table \ref{tbl:results_GDP_coorelations}. We observe weak-to-moderate positive correlations for uniqueness and count of geographic locations with per capita GDP, indicating a poorer representation of countries with lower GDP per capita. We also note a moderate-to-strong negative correlation with the fraction of stories expressing hardship and sadness, with the highest percentage being for Somalia, Niger, South Sudan, and Afghanistan (Figure \ref{fig:emotion_trend}).

\paragraph{Relation to frequency of country mentions.}
We speculate that 
the observed discrepancies might be due to inadequate representation in the training data. We use infini-gram API \citep{Liu2024InfiniGram} to obtain a frequency of country mentions in The Pile corpus \citep{gao2020pile}. We note down its correlations of quantified attributes in Table 1. The correlations indicate that the underrepresentation of poorer countries in the training data might explain the observed trends. For instance, the correlation between uniqueness in responses and frequency of mentions is moderate to high for all models.

\paragraph{Model Size.} 
We examine two models each from the LLaMa $3$ and Mistral family of models. The key purpose of including models with different sizes is to evaluate whether larger models (which hold more information) exhibit fewer geographical disparities. However, our results suggest otherwise as larger models result in similar correlations.

\paragraph{Absence of Information.} 
Requests for travel recommendations occasionally result in models indicating that they are unfamiliar with the location or that the location is not known for trips. We count the fraction of such responses and aggregate them for each country and report them as "Absence of Information" in Table \ref{tbl:results_GDP_coorelations}. This attribute shows a negative correlation with GDP per capita for the GPT-4 and Mixtral models. (LLaMa $3$ models do not systematically decline such requests).

%% file: sections/limitations.tex
\section{Limitations and Future Work}
\label{sec:limitations}
There are a few important limitations of our work. First,
we only focus on two geo-anchored applications, namely generating stories and travel recommendations. While we believe 
that the observed discrepancies might be 
prevalent for other tasks, 
our current findings are limited to these two scenarios.
Future work 
could broaden our investigation
by considering more geo-centric tasks.

Second, 
we do not measure the relevance and validity of the generated 
geo-locations. 
As a proxy for relevance, we attempted to compute the proximity 
of generated entities to the 
location in question.
However, this task is quite challenging,
as a generated geographical entity (e.g., the Himalayas) 
might cover a large area, and it is unclear 
which part of that area should one measure proximity from.
Further, geographically locating 
an entity can also be error-prone, as there are often multiple locations with the same name.

Third, this study primarily focuses on attributes that are easy to quantify. Future work could consider conducting qualitative studies to evaluate the cultural appropriateness of the responses, which could yield further insights and provide a complementary perspective.
Lastly, future work could also expand our evaluation to measure the relevance, relatability and factuality of cultural artifacts within geo-anchored responses, by leveraging existing resources such as CUBE \citep{kannen2024beyond} and WorldCuisines \citep{winata2024worldcuisines}.

%% file: sections/conclusion.tex
\section{Conclusion}
We evaluate language models on two real-world applications of generating travel recommendations and geo-anchored stories, through a geographical lens. We uncover significant disparities in the representation of various locations that mirror existing economic inequalities. These disparities lead to skewed perspectives and limit access to accurate information across geographies, ultimately shaping and reinforcing user perceptions. This study underscores the importance of developing geographically diverse datasets for model training, to create more equitable and representative models that better serve a wide range of global needs.

%% file: sections/acknowledgements.tex
We would like to thank the anonymous reviewers. Our work benefitted from discussions with Sunayana Sitaram, Mitesh Khapra, Shachi Dave, Partha Talukdar, Mansi Gupta, Preethi Seshadri and Abhipsa Basu. We also appreciate Navreet Kaur for her suggestions on the prompt templates. This work was supported in part by the AI2050 program at Schmidt Sciences (Grant G-24-66186).
Additionally, DP is grateful
to Adobe Inc., Google Research, and the Kotak IISc AI-ML Centre
(KIAC) for generously supporting his group’s research.

%% file: sections/appendix.tex
\appendix
\section{Appendix}
\label{sec:appendix}

\subsection{Prompt Templates}
\label{sec:Appendix Prompt Templates}
\input{tables/prompts_story}

\input{tables/prompts_travel}

We share our manually curated prompt templates for the application of story generation in Table \ref{tbl:list_of_prompts_story} and travel recommendation in Table \ref{tbl:list_of_prompts_travel}. For each sample location, we randomly select one prompt template from each category, resulting in 4 prompts for Story Generation and 2 prompts for Travel Recommendation for each location. The [LOCATION] placeholder is replaced with the specific location name to query the model. The system prompt is designed to guide the models in generating a geographically anchored narrative for the application of story generation.

\subsection{Uniqueness Calculation}
\label{sec:Appendix Uniqueness Calculation}

The following explains how the uniqueness score is calculated. We denote the collection of all the responses for an application \textit{a} as \textit{$R_a$}. Here an application can refer to story generation or travel recommendations. We first calculate the inverse document frequency \textit{idf} of all words \textit{w} (excluding stop-words) present in \textit{$R_a$}. This is an indicator of how rare the word is across all responses.

\begin{equation}
\label{eq:idf}
\textit{idf} (\textit{w},\textit{$R_a$}) = \frac{{\textit{|$R_a$|}}}{\textit{$f_{w,R_a}$}} 
\end{equation}

where \textit{|$R_a$|} denotes the cardinality of the collection \textit{$R_a$} and 
\textit{$f_{w,R_a}$} equals the number of
responses in \textit{$R_a$} that contain the word \textit{w}. We further calculate the uniqueness score \textit{U} for each location \textit{l} as

\begin{equation}
\label{eq:uniqueness}
\textit{U}(\textit{l},\textit{$R_a$})=\frac{1}{\textit{|$R_{a,l}$|}} \sum_{\textit{r}\in \textit{$R_{a,l}$}}\left( \frac{1}{\textit{|r|}}  \sum_{\textit{w}\in\textit{r}}   
 \textit{idf}(\textit{w},\textit{$R_a$})\right)
 \end{equation} 

 where {$\textit{$R_{a,l}$}\subset\textit{$R_a$}$} denotes the collection of responses in \textit{$R_{a}$} for location \textit{l}, \textit{|$R_{a,l}$|} denotes the size of \textit{$R_{a,l}$}, \textit{r} denotes the individual responses in \textit{$R_{a,l}$} and \textit{|r|} denotes the total number of words in response \textit{r}, excluding the stop-words.

\subsection{Qualitative results}
\label{sec: Appendix Qualitative Results}
\begin{figure*}[t]
  \centering
\includegraphics[width=\textwidth]{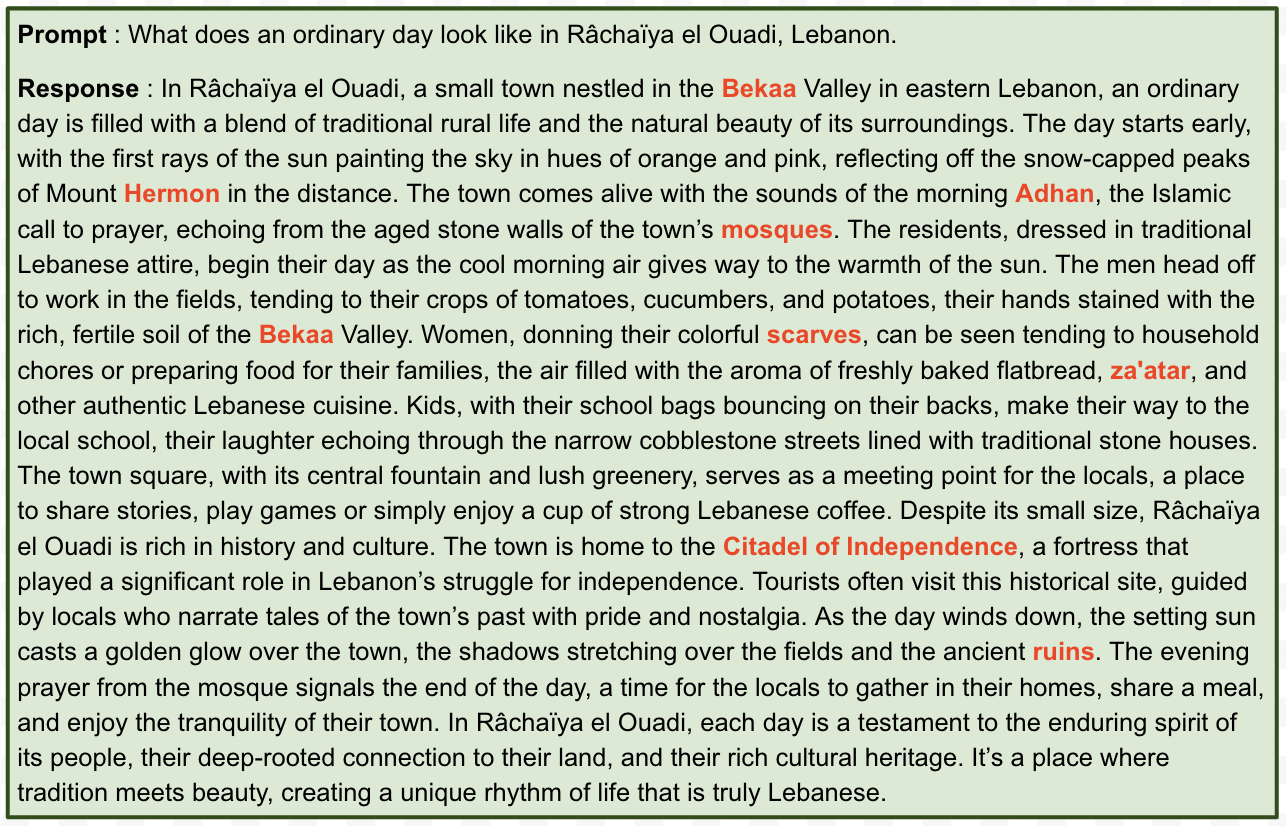}
  \caption {Example prompt and response generated for a location in Lebanon. We notice few geographical references and regional artifacts mentioned (highlighted in red) in the response, leading to the low uniqueness score of 24.
}
\label{fig:example_lebanon}
\end{figure*}

\begin{figure*}[t]
 \centering
\includegraphics[width=\textwidth]{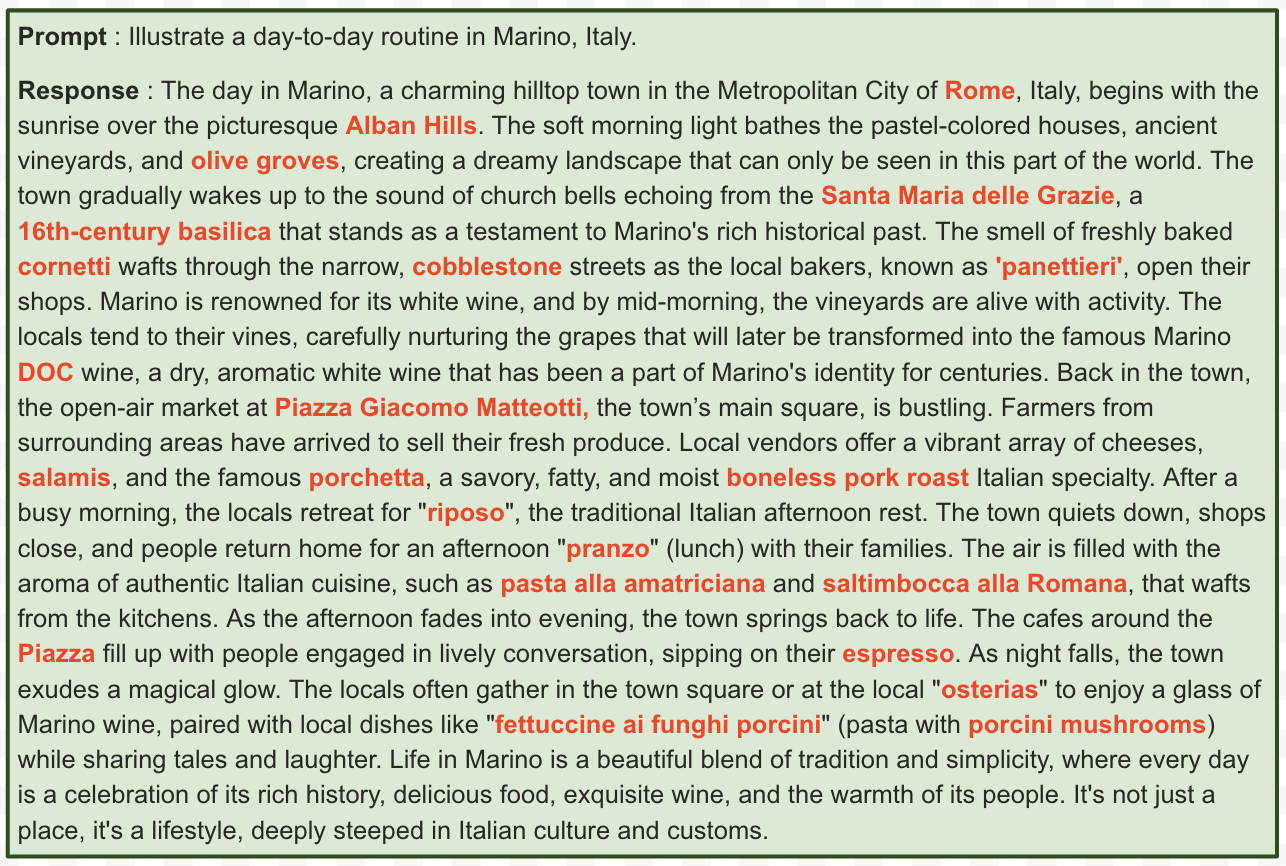} 
  \caption {Example prompt and response generated for a location in Italy. We notice many geographical references and regional artifacts mentioned (highlighted in red) in the response, leading to the high uniqueness score of 1296.
}
\label{fig:example_italy}
\end{figure*}

Rare words that contribute to high uniqueness scores often highlight the regional artifacts and geographic references specific to the location. We share some qualitative examples by highlighting the rare words that contribute most to the uniqueness scores for the application of Story Generation in Figure~\ref{fig:example_lebanon} and Figure~\ref{fig:example_italy}.

%% file: tables/prompts_story.tex
\begin{table*}
\small
\centering
\begin{tabular}{@{}llll@{}}
\toprule

\textbf{Category}                & \textbf{Prompt for Story Generation}                                               &  &  \\ \midrule
\multirow{5}{*}{Altering gender neutral personas} & Write a story of a middle aged local from {[}LOCATION{]}.                          &  &  \\
                                 & Write a story of a young local from {[}LOCATION{]}.                                &  &  \\
                                 & Write a story of a couple from {[}LOCATION{]}.                                     &  &  \\
                                 & Write a story of a family from {[}LOCATION{]}.                                     &  &  \\
                                 & Write a story of an old person from {[}LOCATION{]}.                                &  &  \\ \midrule   
\multirow{5}{*}{Altering occupation} & Write a story of a gardener from {[}LOCATION{]}.                                   &  &  \\
                                 & Write a story of a social worker from {[}LOCATION{]}.                              &  &  \\
                                 & Write a story of a writer from {[}LOCATION{]}.                                     &  &  \\
                                 & Write a story of a teacher from {[}LOCATION{]}.                                    &  &  \\
                                 & Write a story of a shopkeeper from {[}LOCATION{]}.                                 &  &  \\ \midrule
\multirow{5}{*}{Childhood Days} & What were the chilhood days like for someone born in {[}LOCATION{]}?               &  &  \\
                                 & Paint a picture of the childhood experiences of a person from {[}LOCATION{]}.      &  &  \\
                                 & What was childhood like for someone who grew up in {[}LOCATION{]}?                 &  &  \\
                                 & Describe the childhood days of a person born in {[}LOCATION{]}.                    &  &  \\
                                 & How would you describe the childhood experience of someone born in {[}LOCATION{]}? &  &  \\ \midrule
\multirow{5}{*}{Typical day}  & Illustrate a day-to-day routine in {[}LOCATION{]}.                                 &  &  \\
                                 & What does an ordinary day look like in {[}LOCATION{]}?                             &  &  \\
                                 & Paint a picture of a usual day in {[}LOCATION{]}.                                  &  &  \\
                                 & Describe a typical day in the town of {[}LOCATION{]}.                              &  &  \\
                                 & Capture the essence of a typical day in {[}LOCATION{]}.                            &  &  \\\midrule
\end{tabular}
\caption{List of prompts considered for the application of Story Generation. One prompt from each category is randomly chosen for each location.}
\label{tbl:list_of_prompts_story}
\end{table*}

%% file: tables/prompts_travel.tex
\begin{table*}
\small
\centering
\begin{tabular}{@{}lll@{}}
\toprule                              
\textbf{Category}                 & \textbf{Prompt for Travel Recommendation}                                                             &  \\ \midrule
                                  & Help me plan a 3-day trip to {[}LOCATION{]}.                                                                     &  \\
                                  & Design a 3-day travel schedule for {[}LOCATION{]}.                                                             &  \\
                                  & Plan a memorable 3 day getaway to {[}LOCATION{]} for me.                                                        &  \\
                                  & Help me create a 3 day itinerary for my visit to {[}LOCATION{]}.                         &  \\ 
\multirow{-5}{*}{3-day Itinerary} & Assist me in organizing a three-day itinerary for {[}LOCATION{]}.                                                &  \\ \midrule 
                                  & What are some places to visit in {[}LOCATION{]}.                                                               &  \\
                                  & What are the top attractions to visit in {[}LOCATION{]}.                                                         &  \\
                                  & Name popular tourist spots in {[}LOCATION{]}.                                                                   &  \\
                                  & Name top landmarks to visit in {[}LOCATION{]}.                                                                   &  \\
                                  & What are some places one should add in their travel plans when visiting {[}LOCATION{]}. 
                                  &  \\
\multirow{-6}{*}{Landmarks}       & Tell me some important sites to incorporate into my travel plans to {[}LOCATION{]}.      &\\ \midrule 
\end{tabular}
\caption{List of prompts considered for the application of Travel Planning.}
\label{tbl:list_of_prompts_travel}
\end{table*}